\documentclass[11pt,a4paper]{article}
\usepackage[hyperref]{acl2020}
\usepackage{times}
\usepackage{latexsym}
\usepackage{enumitem}
\usepackage{amsmath,amssymb,amsfonts}
\usepackage{multirow}
\usepackage{adjustbox}

\usepackage[utf8]{inputenc}
\usepackage[T5]{fontenc}

\usepackage{microtype}

\aclfinalcopy % Uncomment this line for the final submission
\title{Overview of the VLSP 2023 - ComOM Shared Task: A Data Challenge for Comparative Opinion Mining from Vietnamese Product Reviews}

\author{
    \textbf{Hoang-Quynh Le, Duy-Cat Can, Khanh-Vinh Nguyen and Mai-Vu Tran} \\ 
    VNU University of Engineering and Technology, Hanoi, Vietnam.\\
    {\tt\{lhquynh, catcd, vutm, vinhnk\}@vnu.edu.vn}\\
}

\date{}

\begin{document}
\maketitle
\begin{abstract}
This paper presents a comprehensive overview of the Comparative Opinion Mining from Vietnamese Product Reviews shared task (ComOM), held as part of the 10$^{th}$ International Workshop on Vietnamese Language and Speech Processing (VLSP 2023). The primary objective of this shared task is to advance the field of natural language processing by developing techniques that proficiently extract comparative opinions from Vietnamese product reviews. Participants are challenged to propose models that adeptly extract a comparative ``quintuple'' from a comparative sentence, encompassing Subject, Object, Aspect, Predicate, and Comparison Type Label.
We construct a human-annotated dataset comprising $120$ documents, encompassing $7427$ non-comparative sentences and $2468$ comparisons within $1798$ sentences. Participating models undergo evaluation and ranking based on the Exact match macro-averaged quintuple F1 score.

\end{abstract}

\section{Task Description}
Product reviews contain valuable information, reflecting users' perspectives on diverse aspects of products, including nuanced comparisons between various products/product lines/aspects, etc. Deciphering these comparative opinions in product reviews holds significant importance for both manufacturers and consumers. For manufacturers, it provides a lens through which they can glean insights into the strengths and weaknesses of their products relative to competitors. Simultaneously, consumers can leverage these comparative insights to make more informed purchasing decisions. In light of this imperative, we introduce the ``Comparative Opinion Mining from Vietnamese Product Reviews'' (ComOM) shared task, aimed at streamlining and enhancing this crucial analytical process.

The primary aim of the ComOM shared task is to foster the development of natural language processing models adept at discerning comparative opinions within product reviews. Embedded within each review are sentences that intricately express opinions on various aspects, showing comparisons in multifaceted ways. Participants are challenged to propose models with the capability to extract a specific set of information, termed as a ``quintuple'', from these nuanced comparative sentences:

\begin{itemize}
    \item \textbf{Subject} (\texttt{S}): The entity that is the subject of the comparison. % (e.g., a particular product model).
    This could refer to a specific product, a model within a product line, or any other entity under evaluation.
    
    \item \textbf{Object} (\texttt{O}): The entity that is compared to the subject (e.g., another product model, a general reference, or a baseline for comparison).
    It represents the point of reference against which the subject is evaluated.
    
    \item \textbf{Aspect} (\texttt{A}): The word or phrase about the feature or attribute % of the subject and object
    that is being compared. % (e.g., \textit{battery life}, \textit{camera quality}, \textit{performance}, etc.).
    Aspects could include specific product features such as \textit{battery life}, \textit{camera quality}, \textit{performance}, or any other relevant attribute.
    
    \item \textbf{Predicate} (\texttt{P}): The comparative word or phrase expressing the nature of the comparison. This could involve terms such as \textit{``better than''}, \textit{``worse than''}, \textit{``equal to''}, or any other comparative expression that indicates the relationship between the subject and object.
    
    \item \textbf{Comparison type label} (\texttt{L}) % can be one of these $8$ following labels: 
    indicates the type of comparison made. The label sets include $8$ comparison labels, which represent the sentiment and semantics of comparisons:
    \begin{itemize}
        \item Gradable comparisons (e.g., \textit{``better''}, \textit{``smaller''}): \texttt{COM+} for positive, \texttt{COM-} for negative and \texttt{COM} for other gradable comparisons.
        \item Superlative comparisons (e.g., \textit{``best''}, \textit{``smallest''}): \texttt{SUP+} for positive, \texttt{SUP-} for negative and \texttt{SUP} for other superlative comparisons.
        \item Equal comparison (e.g., \textit{``same''}, \textit{``as good as''}): label \texttt{EQL} is used to annotate no significant difference.%, do not specify positive or negative polarity.
        \item Non-gradable comparison (e.g., \textit{``different from''}, \textit{``unlike''}): label \texttt{DIF} is used to annotate different comparisons.%, do not specify positive or negative polarity.
    \end{itemize}
\end{itemize}

The field of comparative opinion mining in English text has been the subject of numerous studies dedicated to corpus construction and model development \citet{jindal2006mining, kessler2014corpus, younis2020applying, liu2021comparative}. The COQE corpus by \cite{liu2021comparative} stands out as the most recent English comparative corpus, introducing a refined comparative quintuple concept that incorporates comparative preferences and encompasses sentences with multiple comparisons.
Conversely, studies on this problem for Vietnamese are still in their early stages, marked by initial achievements. Published research has predominantly gathered comments or opinions from social media, e-commerce websites, and review platforms, yet it has yet to focus explicitly on comparative opinion mining ~\cite{kieu2010sentiment, nam2014domain, le2016hotel, van2018uit}. The only existing Vietnamese comparative dataset, found in the work of \citet{bach2015mining}, is relatively basic, permitting each sentence to contain only one comparison and annotated solely with the comparative element, lacking the ability to capture comparative semantic and sentiment information.
The absence of a comprehensive benchmarking dataset has constrained the comparison of various techniques for Vietnamese. The ComOM shared task is instituted to offer researchers an opportunity to propose, assess, and advance their research, fostering the growth of Comparative Opinion Mining studies for the Vietnamese language.

The remainder of the paper is organized as follows: Section 2 gives a detailed description of the task data. The next section describes the evaluation metrics and baseline models. Section 4 describes the task results. Finally, Section 5 concludes the paper.

\section{Task Data}
\subsection{Data Creation}
To construct the corpus for the Comparative Opinion Mining from Vietnamese Product Reviews (ComOM) shared task, referred to as the VCOM corpus, a systematic approach was employed to collect and annotate relevant data.
We gathered reviews from prominent Vietnamese review websites and newspapers, such as VnExpress\footnote{\url{http://vnexpress.net/so-hoa/san-pham}} and Tinhte\footnote{\url{http://tinhte.vn}}. 
To ensure the quality and relevance of the data, an initial round of rough filtering was conducted automatically to exclude overly short or irrelevant documents.
Subsequently, a more refined filtering process was applied manually to guarantee the diversity and richness of the corpus.

The annotation task was undertaken by six proficient native speakers who underwent specialized training to ensure consistency and accuracy in the annotation process. The training period spanned several weeks, equipping annotators with a thorough understanding of the task and the annotation guidelines.

A comprehensive annotation guideline, containing detailed instructions and examples, was provided to the annotators to standardize the annotation process. Any discrepancies between the annotators' annotations and those of the authors were meticulously discussed and resolved to maintain annotation consistency.

The annotation process itself extended over a two-month period, during which each example underwent independent annotation by two annotators.
The inter-annotator agreement rate reached $81\%$, indicating a substantial level of agreement.
In cases where discrepancies persisted, annotators were encouraged to provide additional insights, particularly in challenging instances. These cases were systematically discussed with one of the authors, and the feedback received played a crucial role in refining the annotation guidelines for improved clarity and consistency.

\subsection{Data statistics}
%@Cat: Check ho chi so lieu dung chua
% TODO check lại hết số liệu và bổ sung P average length
Table~\ref{tab:corpusstat} shows the main statistics of VCOM corpus.
The corpus is divided into three sub-sets: 
% training set ($60$ documents with $1089$ comparisons), 
% development set ($24$ documents with $471$ comparisons), 
% and test set ($36$ documents with $908$ comparisons).
a training set comprising $60$ documents with $1089$ comparisons, a development set with $24$ documents and $471$ comparisons, and a test set containing $36$ documents with $908$ comparisons.

\begin{table}[!ht]
\centering
% \resizebox{\linewidth}{!}{%
\begin{tabular}{ll}
\hline
 & \textbf{Statistic} \\ \hline
Total documents & $120$ documents \\
Total sentences & $9225$ sentences \\
Non-comparative sentences & $7427$ sentences \\
Comparative sentences & $1798$ sentences \\
% Total comparisons & $2468$ comparisons \\
Multi-comparison sentences & $495$ sentences \\
Total comparisons & $2468$ \\
Comparisons per sentences & $1.37$ \\ \hline
Subjects & $2169$ entities \\
Subject average length & $3.14$ tokens \\
Objects & $1455$ entities \\
Object average length & $3.58$ tokens \\
Aspects & $2068$ entities \\
Aspect average length & $3.62$ tokens \\
Predicate & $2468$ entities \\
Predicate average length & $?.??$ tokens \\ \hline
\end{tabular}%
% }
\caption{VCOM corpus main statistics.}
\label{tab:corpusstat}
%File: Corpus_stat.tng
\end{table}

Given the compilation of VCOM from diverse sources featuring distinct writing styles and characteristics, it exhibits a broad spectrum of linguistic diversity.
Notably, the Out-of-Vocabulary (OOV) rates for all elements in the test set are significantly elevated compared to the training and development sets. Specifically, the OOV rates for the \texttt{Subject}, \texttt{Object}, \texttt{Aspect}, and \texttt{Predicate} stand at $35.59\%$, $37.76\%$, $29.17\%$, and $25.42\%$, respectively.

\begin{figure}[!ht]
    \centering
    \includegraphics[width=0.6\linewidth]{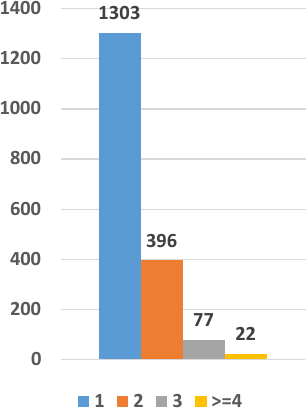}
    \caption{Multi-comparison statistic.}
    \label{fig:multicomparisons}
\end{figure}

One of the complexities characteristic of VCOM corpus is that a sentence can contain multiple comparative quintuples. On average, each comparative sentence contains $1.4$ quintuples. Figure~\ref{fig:multicomparisons} provides a detailed illustration of sentences with one or more than one quintuple.

\begin{figure}[!ht]
    \centering
    \includegraphics[width=0.7\linewidth]{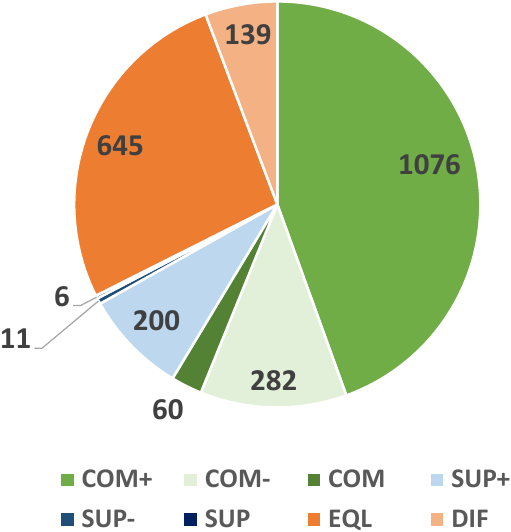}
    \caption{Comparison type labels statistic.}
    \label{fig:labelstat}
\end{figure}

Figure~\ref{fig:labelstat} further illustrates the distribution of the $8$ comparison labels within the VCOM corpus.
While all comparative labels are present in all subsets, there exists a significant imbalance among these labels.
The \texttt{COM+} label dominates, while the \texttt{SUP} and \texttt{SUP-} labels are very limited in quantity. Surprisingly, despite a large number of \texttt{COM+} instances, \texttt{COM-} is substantially fewer, similar to the disparity between \texttt{SUP+} and \texttt{SUP-}.

\subsection{Data Format}
Participants were provided with labeled training and development sets. % for the Comparative Opinion Mining from Vietnamese Product Reviews (ComOM) shared task.
The public leaderboard results were reported based on the development sets, providing participants with a reference for model performance assessment.

The dataset is organized in a structured format to ensure clarity and consistency.
Each document is presented in a plain text (\texttt{txt}) file, with each document further segmented into individual sentences.
Sentences that feature comparisons are meticulously paired with corresponding sets of comparison quintuples.%, offering a comprehensive dataset for training and evaluation.

\begin{figure*}[t]
\begin{minipage}[t]{\textwidth}
    % A line without background color
    \hspace{1em} Below is an example of the provided data format:
    
    % The paragraph with background color
    \colorbox{gray!20}{% Adjust the color and intensity as needed
        \begin{minipage}{\textwidth}
            \texttt{train\_0007.txt\\
            ...\\
            Trọng lượng iPhone 11 Pro sẽ nặng hơn Xs và iPhone 11 Pro Max sẽ có trọng lượng nặng nhất.	Trọng lượng iPhone 11 Pro sẽ nặng hơn Xs và iPhone 11 Pro Max sẽ có trọng lượng nặng nhất .\\
            \{"subject": ["3\&\&iPhone", "4\&\&11", "5\&\&Pro"], "object": ["9\&\&Xs"], "aspect": ["1\&\&Trọng", "2\&\&lượng"], "predicate": ["7\&\&nặng", "8\&\&hơn"], "label": "COM+"\}\\
            \{"subject": ["11\&\&iPhone", "12\&\&11", "13\&\&Pro", "14\&\&Max"], \ \ \ \ \ "object": [], "aspect": ["17\&\&trọng", "18\&\&lượng"], "predicate": ["19\&\&nặng", "20\&\&nhất"], "label": "SUP+"\}\\
            ...}
        \end{minipage}%
    }
\end{minipage}
\end{figure*}

Each comparative sentence and its associated comparison quintuples are structured as follows:
\begin{itemize}
    \item \textbf{Sentence}: The textual content of the sentence. Original text and tokenized text are provided in one line, separated by a tab (`\texttt{\textbackslash t}') character.
    
    \item \textbf{Comparison quintuples}: Each comparison is represented as a quintuple including four comparison elements extracted from the comparative sentence and a comparison label, encoded in JSON format. Each line corresponds to one quintuple.
    The comparison elements are distinctly represented as lists, in the format: \texttt{order\_in\_the\_sentence\&\&word}.
\end{itemize}

%@Cat DONE: them giup chi cai vi du định dạng đẹp đẹp
% đặt phía trên
% Below is an example of the provided data format:
% Input: Bên cạnh đó, việc bổ sung cải tiến SoC mới đã giúp hiệu suất Galaxy S23 Ultra vượt trội hơn Galaxy Z Fold 4 với chip Snapdragon 8 Gen 1.
% Output: \{"subject": ["16\&\&Galaxy", "17\&\&S23", "18\&\&Ultra"], "object": ["22\&\&Galaxy", "23\&\&Z", "24\&\&Fold", "25\&\&4"], "aspect": ["14\&\&hiệu", "15\&\&suất"], "predicate": ["19\&\&vượt", "20\&\&trội", "21\&\&hơn"], "label": "COM+"\}

The format of the provided test set mirrors that of the training and development sets, with the exception that it lacks manually created annotations. The final evaluation was carried out using this test set.

For submissions, participants were required to adhere to a file format identical to that of the training file. Each input corresponded to an output file with an identical filename.
To streamline the submission process, all output files had to be compressed within a \texttt{.zip} file, containing only files (without nested folders). It is crucial to ensure that the total number of sentences in each output file matches the number of sentences in the corresponding testing input.

\section{Evaluation Metrics and Baseline Model}
\subsection{Evaluation Metrics}
We use three matching strategies to evaluate the comparative element extraction:

\begin{itemize}
  \item \textit{Exact Match (}\texttt{E}\textit{)}: The entire extracted element must precisely match the ground truth.
  \item \textit{Proportional Match (}\texttt{P}\textit{)}: We consider the proportion of matched words in the extracted element compared to the ground truth.
  \item \textit{Binary Match (}\texttt{B}\textit{)}: At least one word in the extracted element overlaps with the ground truth.
\end{itemize}

A diverse range of evaluation metrics is applied to assess the effectiveness of comparative opinion models. Precision~(\texttt{P}), Recall~(\texttt{R}), and F1 score~(\texttt{F1}) served as the main measures to gauge the participant's performance.

In assessing the comparative elements extraction (\texttt{CEE}), we calculate Precision, Recall, and F1 score for each individual element (Subject, Object, Aspect, Predicate, and comparison type label), along with their Micro-averaged and Macro-averaged scores.

The evaluation of quintuples involves two types of tuples: 
Tuple of Four (\texttt{T4}, unconsidered comparison type label) and Tuple of Five (\texttt{T5}, considered comparison type label), based solely on exact match and binary match.
    \begin{itemize}
        \item To evaluate \texttt{T4} results, we employ Precision, Recall, and F1. 
        \item For assessing \texttt{T5} results, Precision, Recall, and F1 are utilized for each label, and their Micro-average and Macro-average scores.
    \end{itemize}

The total number of metrics is $120$, and the official evaluation metric is the \textbf{exact match macro-averaged quintuple (\texttt{T5}) F1 score}. Participants will be ranked based on their model performance using this evaluation metric on the test set. 

\paragraph{Evaluation system:} The evaluation was performed on the AIhub\footnote{\url{http://aihub.ml/}} platform. 
The development set is used for evaluation on the public test leaderboard, assisting teams during the experimentation and tuning phases.
Each team was allowed to submit a maximum of $10$ submissions per day.
The maximum number of submissions per user for the public phase is $160$ (~$16$ days).
The test set is employed for evaluation on the private test leaderboard.
The maximum number of submissions per user for the private phase is $5$ in total. 

\subsection{Baseline Model}

\begin{figure*}
    \centering
    \includegraphics[width=\linewidth]{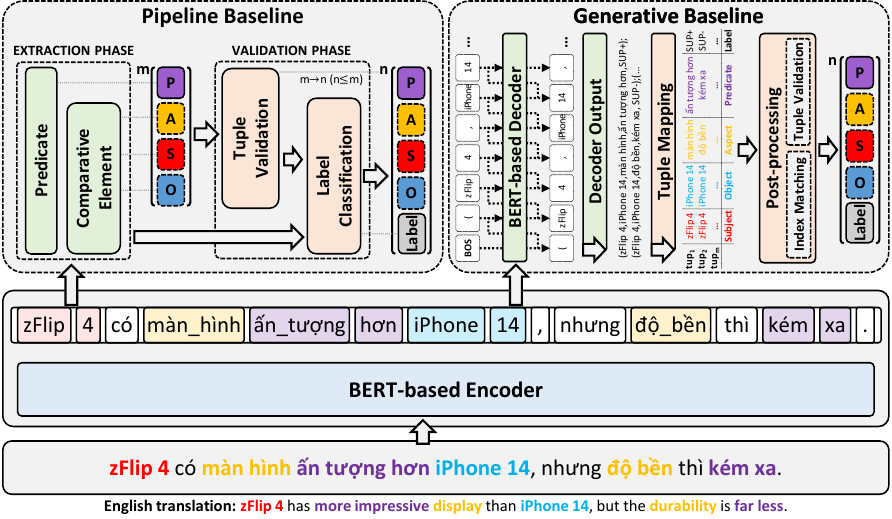}
    \caption{Overall architecture of baseline model}
    \label{fig:overall_architecture}
\end{figure*}

The committee provided a baseline model as the benchmark for the shared task, which is illustrated in Figure~\ref{fig:overall_architecture}. The proposed model consists of an encoding phase, where each sentence is encoded using a BERT-based encoder. The output of the encoding phase is then passed to two different baseline models: the Pipeline baseline and the Generative baseline.

\subsubsection{Pipeline Baseline}
The Pipeline baseline model comprises two distinct phases: extraction and validation.

\paragraph{Extraction Phase:}
The extraction phase involves predicate extraction and the extraction of comparative elements. The sequence of vector outputs from the encoding phase is passed to an LSTM-CRF model for sequence labeling. Each token is assigned a label, where `\texttt{B}' represents the beginning of a predicate, `\texttt{I}' represents the inside of a predicate, and `\texttt{O}' represents tokens outside any predicate. Predicates are then extracted from these labels.

For each extracted predicate, three similar LSTM-CRF models are applied to extract Aspects (\texttt{A}), Subjects (\texttt{S}), and Objects (\texttt{O}). Utilizing the Descartes product, the result is a set of $m$ quadruples of \texttt{(P,A,S,O)}.

\paragraph{Validation Phase:}
For each quadruple, representations of all elements are concatenated, and this vector is passed through a Softmax classification layer to predict if the quadruple is valid. The representation of a valid quadruple is concatenated with the representation of the sentence and passed to another Softmax classification layer to predict the comparison label. In the end, there are $n$ (where $n \leq m$) comparison quintuples.

\subsubsection{Generative Baseline}
The Generative baseline model is based on the concept of an encoder-decoder model.

\paragraph{Generation Phase:}
The output of the encoding phase is fed into a fine-tuned BERT-based decoder to generate entire comparison quintuples in the format of \texttt{(S,O,A,P,L)}. If a sentence contains more than one comparative opinion, these tuples are separated by a semicolon (`\texttt{;}'). Sentences without any comparative opinion are decoded into a special sequence, e.g., ``\texttt{n/a}''.

\paragraph{Post-processing:}
The generated sequence is then split into distinct quintuples with five elements. Tuples that are missing any element are omitted. Two post-processing steps are employed: index mapping and tuple validation.

In the index mapping, if multiple possible indexes of \texttt{(S,O,A,P,L)} exist, the index closest to \texttt{P} as the predicate is chosen. Tuple validation involves checking the original sentence, removing words that do not exist in the original one.

\begin{table*}[!ht]
\centering
\resizebox{\textwidth}{!}{%
\begin{tabular}{lrrrrrrrrr}
\hline
\multicolumn{1}{c|}{\textbf{Team}} & \multicolumn{1}{c}{\textbf{\begin{tabular}[c]{@{}c@{}}\texttt{E-T5-}\\\texttt{MACRO-}\\\texttt{F1}\end{tabular}}} & \multicolumn{1}{c}{\textbf{\begin{tabular}[c]{@{}c@{}}\texttt{E-T5-}\\\texttt{MACRO-}\\\texttt{P}\end{tabular}}} & \multicolumn{1}{c}{\textbf{\begin{tabular}[c]{@{}c@{}}\texttt{E-T5-}\\\texttt{MACRO-}\\\texttt{R}\end{tabular}}} & \multicolumn{1}{c}{\textbf{\begin{tabular}[c]{@{}c@{}}\texttt{E-T5-}\\\texttt{MICRO-}\\\texttt{F1}\end{tabular}}} & \multicolumn{1}{c}{\textbf{\begin{tabular}[c]{@{}c@{}}\texttt{E-T5-}\\\texttt{MICRO-}\\\texttt{P}\end{tabular}}} & \multicolumn{1}{c}{\textbf{\begin{tabular}[c]{@{}c@{}}\texttt{E-T5-}\\\texttt{MICRO-}\\\texttt{R}\end{tabular}}} & \multicolumn{1}{c}{\textbf{\begin{tabular}[c]{@{}c@{}}\texttt{E-T4-}\\\texttt{}\\\texttt{F1}\end{tabular}}} & \multicolumn{1}{c}{\textbf{\begin{tabular}[c]{@{}c@{}}\texttt{E-CEE-}\\\texttt{MACRO-}\\\texttt{F1}\end{tabular}}} & \multicolumn{1}{c}{\textbf{\begin{tabular}[c]{@{}c@{}}\texttt{E-CEE-}\\\texttt{MICRO-}\\\texttt{F1}\end{tabular}}} \\ \hline
\multicolumn{1}{l|}{ABCD Team} & \textbf{23.73}$^{\texttt{1\ }}$ & \textbf{28.62}$^{\texttt{1\ }}$ & 22.16$^{\texttt{2\ }}$ & \textbf{29.52}$^{\texttt{1\ }}$ & 28.80$^{\texttt{2\ }}$ & 30.29$^{\texttt{2\ }}$ & 31.35$^{\texttt{2\ }}$ & 64.11$^{\texttt{2\ }}$ & 63.31$^{\texttt{2\ }}$ \\ \hline
\multicolumn{1}{l|}{NOWJ 3} & 23.00$^{\texttt{2\ }}$ & 20.21$^{\texttt{3\ }}$ & \textbf{27.18}$^{\texttt{1\ }}$ & 26.84$^{\texttt{3\ }}$ & 22.34$^{\texttt{3\ }}$ & \textbf{33.59}$^{\texttt{1\ }}$ & 29.07$^{\texttt{3\ }}$ & 60.79$^{\texttt{5\ }}$ & 59.89$^{\texttt{5\ }}$ \\ \hline
\multicolumn{1}{l|}{VBD\_NLP} & 21.31$^{\texttt{3\ }}$ & 20.93$^{\texttt{2\ }}$ & 21.99$^{\texttt{3\ }}$ & 29.41$^{\texttt{2\ }}$ & \textbf{29.41}$^{\texttt{1\ }}$ & 29.41$^{\texttt{3\ }}$ & \textbf{31.72}$^{\texttt{1\ }}$ & 63.37$^{\texttt{3\ }}$ & 62.84$^{\texttt{3\ }}$ \\ \hline
\multicolumn{1}{l|}{The challengers} & 11.19$^{\texttt{4\ }}$ & 9.64$^{\texttt{9\ }}$ & 13.75$^{\texttt{4\ }}$ & 20.92$^{\texttt{4\ }}$ & 17.09$^{\texttt{8\ }}$ & 26.98$^{\texttt{4\ }}$ & 23.23$^{\texttt{4\ }}$ & \textbf{66.17}$^{\texttt{1\ }}$ & \textbf{65.45}$^{\texttt{1\ }}$ \\ \hline
\multicolumn{1}{l|}{RTX5000} & 9.97$^{\texttt{5\ }}$ & 9.68$^{\texttt{8\ }}$ & 10.65$^{\texttt{6\ }}$ & 17.78$^{\texttt{8\ }}$ & 16.75$^{\texttt{9\ }}$ & 18.94$^{\texttt{9\ }}$ & 19.22$^{\texttt{9\ }}$ & 52.68$^{\texttt{10}}$ & 51.37$^{\texttt{11}}$ \\ \hline
\multicolumn{1}{l|}{KM-doubleQ} & 9.75$^{\texttt{6\ }}$ & 9.33$^{\texttt{10}}$ & 10.57$^{\texttt{7\ }}$ & 18.05$^{\texttt{7\ }}$ & 16.48$^{\texttt{10}}$ & 19.93$^{\texttt{7\ }}$ & 19.74$^{\texttt{8\ }}$ & 58.26$^{\texttt{8\ }}$ & 57.80$^{\texttt{8\ }}$ \\ \hline
\multicolumn{1}{l|}{3N1M} & 9.56$^{\texttt{7\ }}$ & 10.61$^{\texttt{6\ }}$ & 9.09$^{\texttt{10}}$ & 18.68$^{\texttt{5\ }}$ & 18.98$^{\texttt{7\ }}$ & 18.39$^{\texttt{10}}$ & 23.15$^{\texttt{5\ }}$ & 60.10$^{\texttt{6\ }}$ & 59.03$^{\texttt{6\ }}$ \\ \hline
\multicolumn{1}{l|}{Se7enista} & 9.41$^{\texttt{8\ }}$ & 11.34$^{\texttt{5\ }}$ & 8.11$^{\texttt{13}}$ & 16.30$^{\texttt{9\ }}$ & 19.11$^{\texttt{6\ }}$ & 14.21$^{\texttt{13}}$ & 17.94$^{\texttt{11}}$ & 47.58$^{\texttt{12}}$ & 46.49$^{\texttt{13}}$ \\ \hline
\multicolumn{1}{l|}{\textit{\begin{tabular}[c]{@{}l@{}}Generative \\ baseline\end{tabular}}} & \textit{9.23}$^{\texttt{9\ }}$ & \textit{10.41}$^{\texttt{7\ }}$ & \textit{8.46}$^{\texttt{12}}$ & \textit{18.40}$^{\texttt{6\ }}$ & \textit{20.26}$^{\texttt{5\ }}$ & \textit{16.85}$^{\texttt{12}}$ & \textit{21.53}$^{\texttt{6\ }}$ & \textit{56.17}$^{\texttt{9\ }}$ & \textit{55.51}$^{\texttt{9\ }}$ \\ \hline
\multicolumn{1}{l|}{nam} & 8.39$^{\texttt{10}}$ & 7.25$^{\texttt{12}}$ & 10.40$^{\texttt{8\ }}$ & 15.38$^{\texttt{11}}$ & 12.75$^{\texttt{12}}$ & 19.38$^{\texttt{8\ }}$ & 20.19$^{\texttt{7\ }}$ & 59.43$^{\texttt{7\ }}$ & 58.69$^{\texttt{7\ }}$ \\ \hline
\multicolumn{1}{l|}{Chi Linh} & 7.96$^{\texttt{11}}$ & 8.45$^{\texttt{11}}$ & 8.82$^{\texttt{11}}$ & 15.42$^{\texttt{10}}$ & 13.52$^{\texttt{11}}$ & 17.95$^{\texttt{11}}$ & 19.21$^{\texttt{10}}$ & 51.96$^{\texttt{11}}$ & 51.53$^{\texttt{10}}$ \\ \hline
\multicolumn{1}{l|}{WinNLP} & 7.85$^{\texttt{12}}$ & 5.62$^{\texttt{13}}$ & 13.34$^{\texttt{5\ }}$ & 14.87$^{\texttt{12}}$ & 10.50$^{\texttt{13}}$ & 25.44$^{\texttt{5\ }}$ & 15.96$^{\texttt{13}}$ & 47.12$^{\texttt{13}}$ & 46.50$^{\texttt{12}}$ \\ \hline
\multicolumn{1}{l|}{\textit{\begin{tabular}[c]{@{}l@{}}Pipeline \\ baseline\end{tabular}}} & \textit{6.68}$^{\texttt{13}}$ & \textit{5.25}$^{\texttt{14}}$ & \textit{9.98}$^{\texttt{9\ }}$ & \textit{13.37}$^{\texttt{13}}$ & \textit{9.82}$^{\texttt{14}}$ & \textit{20.93}$^{\texttt{6\ }}$ & \textit{16.96}$^{\texttt{12}}$ & \textit{60.83}$^{\texttt{4\ }}$ & \textit{60.11}$^{\texttt{4\ }}$ \\ \hline
\multicolumn{1}{l|}{Beginners} & 6.27$^{\texttt{14}}$ & 12.38$^{\texttt{4\ }}$ & 4.20$^{\texttt{14}}$ & 11.18$^{\texttt{14}}$ & 22.01$^{\texttt{4\ }}$ & 7.49$^{\texttt{14}}$ & 12.13$^{\texttt{14}}$ & 24.62$^{\texttt{16}}$ & 24.29$^{\texttt{16}}$ \\ \hline
\multicolumn{1}{l|}{E11} & 1.08$^{\texttt{15}}$ & 1.28$^{\texttt{16}}$ & 0.97$^{\texttt{15}}$ & 2.70$^{\texttt{16}}$ & 3.06$^{\texttt{16}}$ & 2.42$^{\texttt{15}}$ & 3.19$^{\texttt{15}}$ & 29.50$^{\texttt{14}}$ & 28.56$^{\texttt{14}}$ \\ \hline
\multicolumn{1}{l|}{otaku} & 1.07$^{\texttt{16}}$ & 1.81$^{\texttt{15}}$ & 0.87$^{\texttt{16}}$ & 2.72$^{\texttt{15}}$ & 3.55$^{\texttt{15}}$ & 2.20$^{\texttt{16}}$ & 3.12$^{\texttt{16}}$ & 28.21$^{\texttt{15}}$ & 27.18$^{\texttt{15}}$ \\ \hline
\multicolumn{1}{l|}{Happy Boy} & 0.79$^{\texttt{17}}$ & 0.87$^{\texttt{17}}$ & 0.78$^{\texttt{17}}$ & 1.60$^{\texttt{17}}$ & 1.67$^{\texttt{17}}$ & 1.54$^{\texttt{17}}$ & 1.72$^{\texttt{17}}$ & 22.71$^{\texttt{17}}$ & 23.16$^{\texttt{17}}$ \\ \hline
\multicolumn{1}{l|}{Aladin} & 0.05$^{\texttt{18}}$ & 0.04$^{\texttt{18}}$ & 0.06$^{\texttt{18}}$ & 0.20$^{\texttt{18}}$ & 0.18$^{\texttt{18}}$ & 0.22$^{\texttt{18}}$ & 0.19$^{\texttt{18}}$ & 13.22$^{\texttt{18}}$ & 13.78$^{\texttt{18}}$ \\ \hline
\multicolumn{10}{r}{\begin{tabular}[c]{@{}r@{}}Results are shown in \%. The highest result in each column is highlighted in bold. \\ The number in the superscript ($^{\texttt{n}}$) indicates the corresponding rank of a score. \\ Baseline results are displayed in italic.\end{tabular}}
\end{tabular}%
}
\caption{Official Results on the Private Test.}
\label{tab:result}
%File: results.tng
\end{table*}

\section{Results}
\subsection{Participant}
There are $48$ registered teams from research groups in domestic universities (HUST, VNU-HUS, VNUHCM-UIT, VNU-UET, HCMUS, HCMUT, PTIT, FTU, etc.), international universities (Minerva University - US, University of Chicago - US, etc.) and industries (Viettel, VinBigData, VNPT-IT, TopCV, VCCorp, etc). 
$20$ teams participated officially by submitting at least $1$ run on the evaluation platform.
%Participant teams can use all possible tools and resources to build models.
Participated teams made a total of $52$ submissions.
Post-challenge panels\footnote{\url{http://aihub.ml/competitions/601}} are now opened on AIHUB for supporting research improvements.

% The proposed models followed two main approaches...
%Quynh cho Vinh tom tat method cua top 5

\subsection{Task Results}
%@Cat: Giai thich ho chi
% An interesting observation is that the pipeline baseline achieved higher results than the generative baseline in the public test, but lower in the private test. 
An interesting observation arises from the comparative evaluation of baseline models. In the public test set, the Pipeline baseline outperformed the Generative baseline, showcasing its effectiveness in the specific characteristics of that dataset. However, in the more diverse private test set, the Generative baseline surpassed the Pipeline baseline. 
This intriguing shift in performance suggests a nuanced interaction between the baseline models and the distinct characteristics of the test sets. The public test set favored the structured approach of the Pipeline baseline, while the Generative baseline demonstrated greater adaptability to the challenges posed by the private test set.

The results of the private test were considered as the official results to rank the team in ComOM shared task. 
Table~\ref{tab:result} shows the results on the exact match macro/micro-averaged Precision, Recall, F1 score of the tuple of five (\texttt{E-T5-MACRO-P/R/F1} and \texttt{E-T5-MICRO-P/R/F1}), exact match F1 score of the tuple of four (\texttt{E-T4-F1}), and exact match macro/micro-averaged F1 score of comparative elements extraction (\texttt{E-CEE-MACRO-F1} and \texttt{E-CEE-MICRO-F1}). 
 The macro-averaged F1 of the tuple of five (\texttt{E-T5-MACRO-F1}) serves as the official evaluation metric for team ranking.
Among the participating teams, $11$ achieved superior performance compared to the pipeline baseline, with $8$ teams surpassing the generative baseline in terms of \texttt{E-T5-MACRO-F1}.

The highest \texttt{E-T5-MACRO-F1} achieved was $23.73\%$, with corresponding \texttt{E-T5-MACRO-P} and \texttt{E-T5-MACRO-R} values at $28.62\%$ and $22.16\%$ respectively. Additionally, the top-performing team secured the best \texttt{E-T5-MICRO-F1} at $29.52\%$.
It is noteworthy that the team ranked third achieved the highest \texttt{E-T4-F1} score at $31.72\%$. Additionally, the team placed fourth and excelled in both \texttt{E-CEE-MACRO-F1} and \texttt{E-CEE-MICRO-F1}, securing impressive scores of $66.17\%$ and $65.45\%$ respectively.

\section{Conclusions}
The VLSP 2023 - ComOM shared task was formulated to propel research development in the domain of comparative opinion mining in Vietnamese. Our objective is to juxtapose diverse approaches in comparative opinion mining, offering a standardized test-bed for forthcoming research endeavors.
The meticulously crafted VCOM dataset is anticipated to yield substantial contributions to related works.
The ComOM shared task garnered attention from the research community, attracting participating teams that employed varied approaches, advanced technologies, and diverse resources. We have archived noteworthy and promising results, establishing valuable benchmarks for future investigations.
In conclusion, we are pleased to affirm that the VLSP 2023 - ComOM shared task unfolded successfully, poised to make significant contributions to the Vietnamese text mining and natural language processing communities.

% \section*{Acknowledgments}

% The acknowledgments should go immediately before the references. Do not number the acknowledgments section.
% Do not include this section when submitting your paper for review.

\bibliography{anthology,acl2020}
\bibliographystyle{acl_natbib}

%\appendix

%\section{Appendices}
%\label{sec:appendix}
%Appendices are material that can be read, and include lemmas, formulas, proofs, and tables that are not critical to the reading and understanding of the paper. 

\end{document}